\documentclass[10pt,journal]{IEEEtran}

\usepackage{setspace}
\usepackage{amsmath}
\usepackage{amssymb}
\usepackage{graphicx}
\usepackage{comment}
\usepackage{amsmath,amssymb} 
\usepackage{color}
\usepackage{subfigure}
\usepackage{booktabs} 
\usepackage{hyperref}
\usepackage{bm}

\usepackage{multirow}
\usepackage{multicol}
\usepackage{hyphenat}
\usepackage{ragged2e}
\usepackage{enumerate}

\usepackage{algorithm}  
 \usepackage{algorithmic}  

\ifCLASSOPTIONcompsoc
  \usepackage[nocompress]{cite}
\else
  \usepackage{cite}
\fi

\graphicspath{{figs/}}

\def\lyu{\textcolor{red}}

\def\mb{\mathbf}

\def\mc{\mathcal}

\def\st{\mbox{s.t. }}
\def\ie{\textit{i.e.}}
\def\wrt{\textit{w.r.t. }}
\def\eg{\textit{e.g.}}
\def\etc{\textit{etc.}}

\begin{document}
%
\title{Overcoming Domain Drift in Online\\Continual Learning}
%
%
%
%

\author{Fan Lyu$^1$,~\IEEEmembership{Member,~IEEE,}
        Daofeng Liu$^2$,
        Linglan Zhao$^3$, 
        Zhang Zhang$^1$,~\IEEEmembership{Member,~IEEE,}
        Fanhua Shang$^4$,\\
        Fuyuan Hu$^2$,~\IEEEmembership{Member,~IEEE,}
        Wei Feng$^{4*}$,~\IEEEmembership{Member,~IEEE,}
        Liang Wang$^1$,~\IEEEmembership{Fellow,~IEEE}
        \thanks{$^1$F. Lyu, Z. Zhang and L. Wang are with the New Laboratory of Pattern Recognition (NLPR), Institute of Automation, Chinese Academy of Sciences, Beijing 100190, China. Contacts:  fan.lyu@cripac.ia.ac.cn, \{zzhang, wangliang\}@nlpr.ia.ac.cn}%
        \thanks{$^2$D. Liu and F. Hu are with the School of Electroic\&Engineering, Suzhou University of Science and Technology, Suzhou, 215000 China. Contacts: \{daofengliu@post, fuyuanhu@mail\}.usts.edu.cn.}
        \thanks{$^3$L. Zhao is with the Youtu Lab, Tencent, Shanghai, 200240 China. Contacts: linglanzhao@tencent.com.}
        \thanks{$^4$F. Shang and W. Feng are with the College of Intelligence and Computing, Tianjin University, Tianjin, 300350 China. Contacts: fhshang@tju.edu.cn, wfeng@ieee.org.} 
        \thanks{*The corresponding author is Wei Feng.}
        }

%
%

\markboth{Journal of \LaTeX\ Class Files,~Vol.~14, No.~8, August~2024}%
{Shell \MakeLowercase{\textit{et al.}}: Bare Demo of IEEEtran.cls for Computer Society Journals}
%



\IEEEtitleabstractindextext{%
\begin{abstract}
  \justifying
  {
  Online Continual Learning (OCL) empowers machine learning models to acquire new knowledge online across a sequence of tasks.
  However, OCL faces a significant challenge: catastrophic forgetting, wherein the model learned in previous tasks is substantially overwritten upon encountering new tasks, leading to a biased forgetting of prior knowledge.
  Moreover, the continual doman drift in sequential learning tasks may entail the gradual displacement of the decision boundaries in the learned feature space, rendering the learned knowledge susceptible to forgetting. 
  To address the above problem, in this paper, we propose a novel rehearsal strategy, termed Drift-Reducing Rehearsal (DRR), to anchor the domain of old tasks and reduce the negative transfer effects.
  First, we propose to select memory for more representative samples guided by constructed centroids in a data stream.
  Then, to keep the model from domain chaos in drifting, a two-level angular cross-task Contrastive Margin Loss (CML) is proposed, to encourage the intra-class and intra-task compactness, and increase the inter-class and inter-task discrepancy.
  Finally, to further suppress the continual domain drift, we present an optional Centorid Distillation Loss (CDL) on the rehearsal memory to anchor the knowledge in feature space for each previous old task.
  Extensive experimental results on four benchmark datasets validate that the proposed DRR can effectively mitigate the continual domain drift and achieve the state-of-the-art (SOTA) performance in OCL.
  }
\end{abstract}

}

\maketitle

\IEEEdisplaynontitleabstractindextext

%
\IEEEpeerreviewmaketitle

\section{Introduction}\label{sec:introduction}

\IEEEPARstart{I}N recent decades, machine learning~\cite{alpaydin2021machine} and deep learning~\cite{he2016deep} algorithms leveraging large-scale datasets have demonstrated comparable capabilities to human intelligence in certain single domains~\cite{lyu2019attend}. 
However, unlike humans, who continuously acquire novel knowledge throughout their lifetimes, previous studies often operate under impractical static scenarios, assuming that the class space remains constant.
Continual Learning (CL)~\cite{MetaSP,tnnls3}, also known as lifelong learning and incremental learning, have been proposed for models to learn continuously from a sequence of tasks over an extended period. This CL setting aligns machine learning more closely with realistic human learning, where new skills are acquired continuously with the introduction of fresh training data. 
CL has been proven effective in various applications, including autopilot systems~\cite{jehanzeb2022efficient}, emotion recognition~\cite{thuseethan2021deep}, human motion prediction~\cite{yasar2021improving}, Internet of Things (IoT)~\cite{liu2021class}, and multi-label recognition~\cite{9859622}.
This paper specifically focuses on exploring more challenging Online Continual Learning (OCL) in a general context with streaming data, where each data point or mini-batch traverses the model only once.

\begin{figure}[t]
	\centering	
  \subfigure[OCL with continual domain drift]{\includegraphics[width=\linewidth]{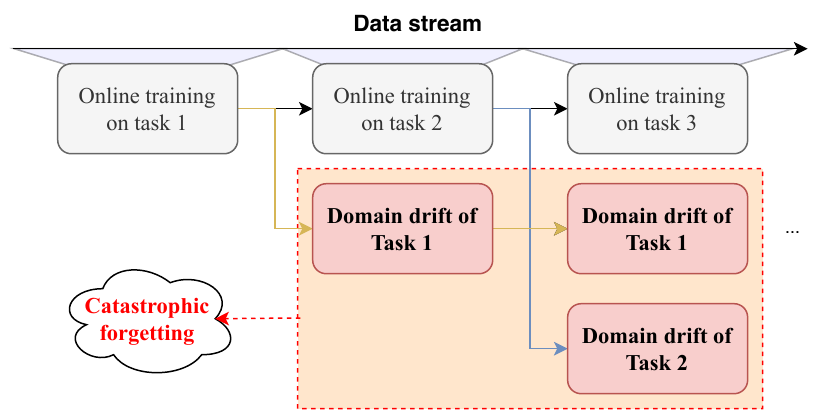}
  \label{fig:1_ocl}
  }
  \subfigure[Continual domain drift in feature space]{\includegraphics[width=\linewidth]{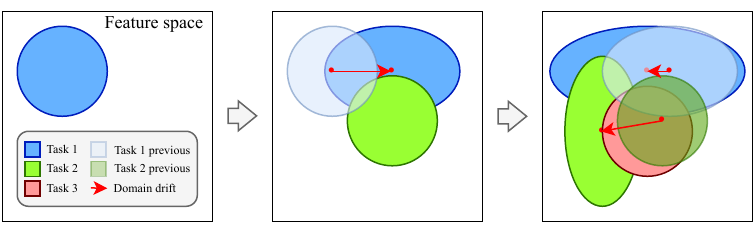}
  \label{fig1:1_drift}
  }
	\caption{
    (a) Task of OCL.
    (b) Continual domain drift in OCL.
    Through continual training, old tasks have their domain distribution drift in an unpredictable direction, and the decision boundaries between old and new tasks could be blurred.
    Best view in color.
	}
  \label{fig:shift}
\end{figure}

Throughout, \textit{catastrophic forgetting}~\cite{french1999catastrophic,kirkpatrick2017overcoming} remains a primary challenge in CL, signifying the risk of overwriting previous knowledge when acquiring new information. This issue becomes even more challenging in the context of OCL, where data is accessed in flow.
The challenge of forgetting is exacerbated, when training the model for new tasks completely excludes previous data.
In recent years, some previous studies have achieved less catastrophic forgetting of old knowledge by finding a balance between old and new tasks in OCL.
Following the categorization in~\cite{de2019continual}, these methods can be classified into categories based on \textit{Regularization}, \textit{Rehearsal}, and \textit{Architecture}. 
Regularization-based methods, exemplified by~\cite{li2016learning,chaudhry2018riemannian,dhar2019learning}, focus on constructing effective regularization terms within the loss function. This type of method doesn't require the storage of previous data, thus safeguarding privacy.
Architecture-based methods are more appropriate for the scenarios where there are no constraint on model size, they identify and freeze task-related parameters while growing new parameter branches to accommodate new knowledge, such as~\cite{mallya2018piggyback,yoon2017lifelong,tnnls2},
Rehearsal-based methods like~\cite{lopez2017gradient,AGEM,guo2019learning} resort to store and re-train a small subset of training samples from old tasks to mitigate catastrophic forgetting. 
The memory size in this case is significantly smaller than that of the entire training set.
In this case, since the memory size is significantly smaller than that of the entire training set, an optimal balance between performance and training cost can be attained.

In addition to the catastrophic forgetting problem in CL, the phenomenon of continual domain drift~\cite{caccia2022new,zenisek2019concept,bhatia2021memstream,tnnls1} in OCL poses more challenge.
As shown in Fig.~\ref{fig:shift}, OCL receives streaming data, and insufficient learning can lead to difficulties in acquiring and retaining knowledge.
Through continual training, the serious data isolation between old and new tasks leads to the trained domains being prone to drift in an undesired direction and even blurs the decision boundaries between old and new tasks, which significantly manifests as the catastrophic forgetting.
\emph{The negative influence of the continual domain drift derives from the lackage of the data from old tasks.}
For regularization-based and architecture-based methods, both of them do not access previous data, where the domain drift remains uncontrollable.
For rehearsal-based methods, which have limited access to previous data, it is challenging to accurately representing the original data distributions using only a subset of the training data from old tasks.
Thus, avoiding the biases introduced by memorizing inappropriate data from previous tasks is crucial for the effectiveness of rehearsal-based methods


%

This paper aims to address the issue of continual domain drift to suppress its negative influence in OCL.
We also build our method on the rehearsal strategy, and our motivation is in two aspects.
First, the negative influence of the continual domain drift derives from the shrinkage of old data via sampling, where the quality of the stored memory buffer is important.
The memory with poor generalization will inevitably worsen the catastrophic forgetting.
Thus, more representative data rather than randomly-sampled data should be stored.
Second, the imbalance of new data and memory data leads to the drift inevitablly toward new tasks, which may blur the decision boundary of old tasks.
This occurs because new tasks provide larger gradients than old tasks in rehearsal-based training.
In this case, when training on both memory and new task data, it is essential to contrain domain drifts and keep a distinc decision boundary.


Motivated by the above considerations, we propose a \emph{Drift-Reducing Rehearsal} (DRR) approach based on the construction of centroids, which selects informative data samples of old tasks as the learned knowledge in memory, comprising two main parts.
First, we propose to store more representative data based on constructed the centroids for each class during a full online phase.
We sample data from the class adaptively via the distance from all centroids.
Moreover, a fixed-size memory buffer will be reallocated when the centroids update with the new accessed data.
Second, we combine the logits of all seen tasks and separate them from each other using a novel cross-task \textit{Contrastive Magin Loss} (CML), in which two levels' angular margins are used to encourage the intra-class/task compactness and the inter-class/task discrepancy.
{The class and task level margins are placed in the angle between any two features through a cross-task contrastive loss.}
CML avoids the vague decision boundary in feature spaces and ensures more effective training on old and new tasks. 
Moreover, we present a new optional \textit{Centroid Distillation Loss} (CDL) on rehearsal memory for old tasks that suppress the continual domain drift by storing the corresponding feature representations of each data point in memory.
We evaluate our proposed DRR method on four popular OCL datasets and the experimental results show our DRR can significantly mitigate the continual domain drift and achieve the SOTA performance.


{Our main contributions are three-fold:
\begin{enumerate}[(1)]
  \item We propose a new Centroid-based Online Selection (COS) strategy to sample more representative samples in streaming rehearsal-based continual learning. The proposed strategy is efficient and suitable for the CL task based on data stream setting.
  \item We propose a novel cross-task Contrastive Margin Loss (CML) for continual learning to further connect each isolation task and set two-level angular margins in the contrastive loss to encourage the intra-class/task compactness and the inter-class/task discrepancy.
  \item We evaluate our methods on four popular OCL datasets and achieve new SOTA performance. All the results show that the proposed method is able to reduce the continual domain drift in OCL and alleviates the catastrophic forgetting problem.
\end{enumerate}

\section{Related Work}\label{sec:rel}

\subsection{Continual Learning and Online Continual Learning}

In contrast to static machine learning~\cite{he2016deep,deng2018visual,lyu2019attend,lyu2020vtgraphnet}, continual learning~\cite{ring1997child,thrun1998lifelong} is designed to mimick human learning that learns new knowledge sequentially.
Because of the learning from sequential tasks, the \textit{catastrophic forgetting} issue~\cite{french1999catastrophic,kirkpatrick2017overcoming} is produced that new knowledge overwrites old knowledge.
Previous solutions to the catastrophic forgetting problem~\cite{french1999catastrophic,kirkpatrick2017overcoming} can be categorized into \textit{regularization-based}, \textit{architecture-based} and \textit{rehearsal-based} methods~\cite{de2019continual}.
Regularization-based methods such as~\cite{li2016learning,chaudhry2018riemannian,dhar2019learning} focus on designing effective regularization terms in the loss function, which restricts the update of important parameters of old tasks.
Architecture-based methods such as~\cite{mallya2018piggyback,yoon2017lifelong} find task-specific paths in neural networks to avoid the new update affecting old tasks.
Some of these methods grow new parameter branches for new knowledge.
Although much progress has been made in regularization-based and architecture-based methods, their performance still has a big gap with that of rehearsal-based methods elaborated in the next subsection.

\subsection{Rehearsal strategy}

Different from regularization-based and architecture-based methods, Rehearsal-based methods consider solving catastrophic forgetting by storing a small-size subset of the old tasks' training data.
First, some methods save the raw data~\cite{Rebuffi2016,lopez2017gradient,AGEM,guo2019learning} and directly retrain the saved data together with the new task training, where the memory is always treated as the constraints.
Then, some methods store the latent features for selected samples (Latent-rehearsal)~\cite{Pellegrini2019}, and build distillation loss to reduce forgetting.
Finally, some methods build generative models such as Generative Adversarial Networks (GANs), to synthesize data from old tasks~\cite{Shen2020,Ven2018,Lesort2018}.
In this way, the knowledge can be saved as parameters rather than raw data or features.

In this paper, we only consider the native rehearsal strategy by storing raw data in OCL. 
The traditional rehearsal methods~\cite{lyu2021multi,AGEM,guo2019learning} take a random selection strategy.
However, randomly sampled data lacks the capacity of representing the whole training set, where the distribution changes significantly with bias.
The shrinkage of the training set makes the retraining of rehearsal prone to continual domain drifts in the feature space.
{Ring buffer~\citen{Chaudhry2019} allows memory to be updated cyclically and only stores the last seen sample.
Mean-of-Feature~\cite{Rebuffi2016} samples the data closest to the mean by calculating the mean of the features of all data in each class.
Gradient-based Sample Selection~\cite{gradient} diversifies the gradients of the samples in the memory buffer.}
However, these previous methods suffer from the following limitations.
(1) They select samples not damage the current new tasks but ignore the future tasks, some non-representative samples are selected.
(2) Some methods rely on storing extra models to evaluate the stored sample, which means more memory is used.
(3) Some methods need to select from the whole training set and thus cannot be applied to streaming data.
(4) Some methods are complex and not efficient in streaming data.
In this paper, we propose a centroid-based online selection strategy, which can be efficiently applied to streaming data in OCL.

\subsection{Contrastive learning}
The goal of contrastive learning is to learn a more discriminative feature representation by learning how to group similar samples together and spread out dissimilar samples~\cite{scl,surveyscl}. 
Specifically, contrastive learning makes positive samples as close to each other as possible and negative samples as far apart as possible by comparing positive samples (similar samples) with negative samples (dissimilar samples). SimCLR~\cite{simclr} propose a framework for self-supervised contrastive learning that leverages a large number of negative samples to learn powerful representations.
Momentum Contrast~\cite{moco} introduce a memory bank and a momentum update strategy to improve contrastive learning.
In the field of continual learning, contrastive learning methods have been shown ability to learn more discriminative representations compared to traditional cross-entropy learning methods~\cite{margincl,scale}. 
They also exhibit better generalization ability and help alleviate the forgetting of previously learned classes in continual learning.
However, existing CL methods using contrastive loss~\cite{cha2021co2l} ignore the continual domain drift problem, and prone to entangle across different tasks and different classes.

\subsection{Margin loss and distillation loss}

Most classification methods use the softmax function as the classifier that maps the logits from a neural network to a probability.
However, the vanilla softmax function is based on the expontional function and has no sensitivity on the target logit and others.
In recent years, some methods add margins to the expontional function to improve the ability of discrimination.
SphereFace~\cite{DBLP:conf/cvpr/LiuWYLRS17} and L-Softmax~\cite{liu2016large-margin} add multiplicative angular margins on the angular space. 
CosFace~\cite{DBLP:conf/cvpr/WangWZJGZL018,DBLP:journals/spl/WangCLL18} and ArcFace~\cite{deng2018arcface} add additive cosine margin and angular margin, respectively. 
This paper proposes a novel cross-task contrastive margin loss inspired by ArcFace. 
Our loss function improves both inter-class/task compactness and intra-class/task discrepancy.
The knowledge distillation~\cite{Hinton2015} is first proposed to transfer knowledge from a teacher network to a student network.
Inspired by the distillation loss, we propose to store the feature representation of memory and build distillation loss between the old and new models on old tasks.



\section{Preliminary: Rehearsal-based OCL}

\renewcommand{\algorithmicrequire}{ \textbf{Procedure}}     

\begin{figure*}
  \centering
  \includegraphics[width=\linewidth]{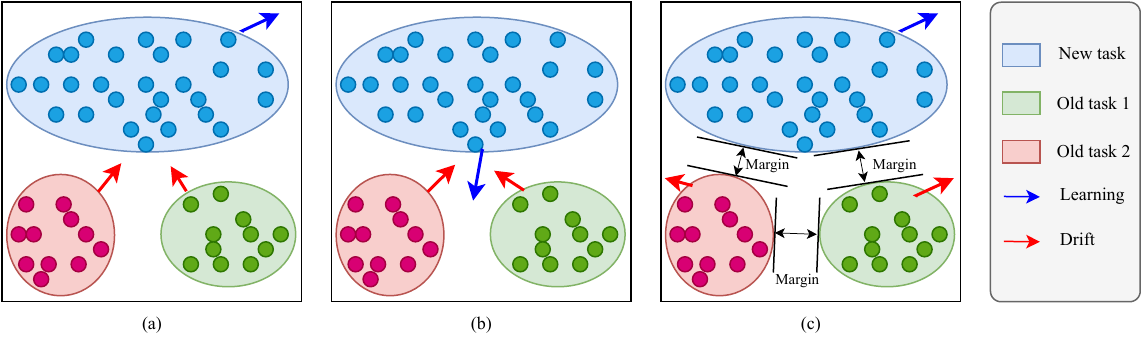}
  \vspace{-20px}
  \caption{
  Drift comparison in rehearsal-based OCL.
  (a) and (b) show the two kinds of domain drift in rehearsal-based CL. Because of the unrepresentative stored data and the gradient bias, the features of old and new tasks are entangled thus causing catastrophic forgetting.
  (c) Our DRR seeks to store representative data via centroids and constrain the drift via the proposed cross-task contrastive margin loss.
  }
  \label{fig:drift}
\end{figure*}

Given $T$ different tasks w.r.t. datasets $\{\mc{D}_1,\cdots,\mc{D}_T\}$. 
For the $t$-th dataset of the corresponding task, $\mc{D}_t=\{(x_{1},y_{1}),\cdots,(x_{N_t},y_{N_t})\}$, where $x_{i}\in\mc{X}_t$ is the $i$-th input data, $y_{i}\in\mc{Y}_t$ is the corresponding label and $N_t$ is the number of samples.
OCL aims to online learn a predictor $f:\mc{X}_k\rightarrow\mc{Y}_k,~k\in\{1,\cdots,t\}$ that predicts all seen tasks in sequence learning at any time $t$.
In the predictor, $\bm{\theta}$ is shared across all seen tasks while $\bm{\theta}_k$ is the task-specific parameter for task $k$. 
Note that in the general online setting, the model has the flexibility to process either individual data points or small-sized mini-batches at each iteration.
In this paper, we focus on batch-level OCL.

The rehearsal-based OCL~\cite{Rebuffi2016,lopez2017gradient,riemer2018learning,AGEM,guo2019learning} builds a memory buffer $\mc{M}_k\subset\mc{D}_k$ with small-size for each previous task $k$, \ie, $|\mc{M}_k|\ll|\mc{D}_k|$. 
Following~\cite{lopez2017gradient}, when training a task $t\in\{2,\cdots,T\}$, for all $\mc{M}_k$ that $k<t$, the rehearsal-based OCL can be formulated as a multi-objective optimization problem: 
\begin{equation}
  \mathop{\min}_{\bm{\theta},\{\bm{\theta}_1,\cdots,\bm{\theta}_t\}}
  \left\{
    \ell(f_{\bm{\theta}},  f_{\bm{\theta}_t}, \mc{D}_t)\right\}
    \cup
    \left\{
    \ell(f_{\bm{\theta}},  f_{\bm{\theta}_{k}},\mc{M}_{k})|\forall k < t
    \right\}
\end{equation}
where $\ell$ is the empirical loss. 
It is difficult to directly solve the above Multi-Objective Optimization (MOO) problem.
In the following, we introduce two main kinds of rehearsal methods for solving the MOO problem, and illustrate the continual domain drift problem.

First, some studies~\cite{lopez2017gradient,AGEM,guo2019learning} design to transform the multi-objective optimization problem to only optimize the new task while constraining the loss rise of old tasks:
\begin{equation}
	\begin{aligned}
    \mathop{\min}_{\bm{\theta},\bm{\theta}_t}\quad&\ell(f_{\bm{\theta}}, f_{\bm{\theta}_t}, \mc{D}_t),\quad\\
		\st\quad&\ell(f_{\bm{\theta}}, f_{\bm{\theta}_k},\mc{M}_k)\le\ell(f_{\bm{\theta}}^{t-1},f^{t-1}_{\bm{\theta}_k}, \mc{M}_k),\quad\forall k<t.
	\end{aligned}
	\label{eq:eps}
\end{equation}
Then, the problem is reduced to find an optimal gradient that approaches the gradient of new task and may not harm any old tasks.
To inspect the increase in old tasks' loss, previous studies \cite{lopez2017gradient,AGEM,guo2019learning} prefer to make the gradient of the current task as the target gradient and compute the angle from the gradients of old tasks to the target gradient. 
Consequently, gradient bias induces rehearsal drift, which in turn causes the features of old tasks to shift towards learning new tasks. This results in feature entanglement and degradation of old task performance, thus exacerbating the catastrophic forgetting (see Fig.~\ref{fig:drift}(a)).
In conclusion, this kind of drift over-emphasizes the current task while ignoring the adaptation of old tasks, which means rigorous forgetting.

\begin{figure*}[t]
	\centering
	\includegraphics[width=\linewidth]{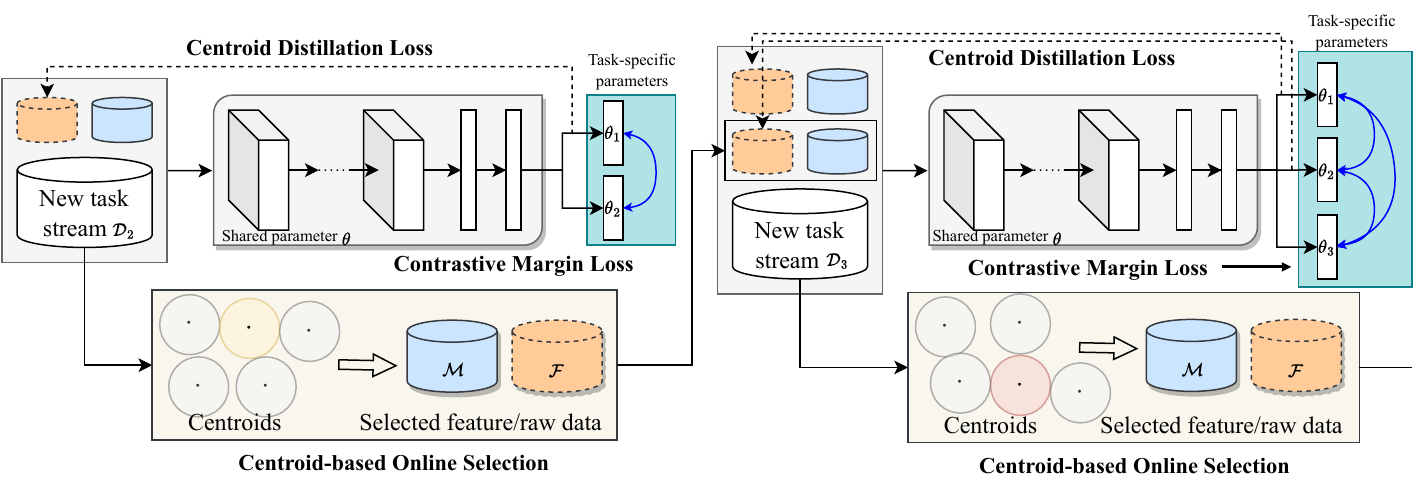}
  \vspace{-20px}
	\caption{
		Training procedure of the proposed DRR in continual learning on a data stream.
		At each step, we store a small number of samples and the corresponding latent representations via centroid-based rehearsal.
		The cross-task margin loss guarantees the intra-class/task compactness and inter-class/task discrepancy.
		The centroid distillation loss helps further to reduce the continual domain drift of the old tasks.
		The dashed elements mean the optional items.
	}
	\label{fig:method}
\end{figure*}

Second, some other methods such as~\cite{chaudhry2019on} ensemble all tasks' objectives as a single-objective problem as follows:
\begin{equation}
	\mathop{\min}_{\bm{\theta},\{\bm{\theta}_1,\cdots,\bm{\theta}_t\}}\quad\alpha_t\ell(f_{\bm{\theta}},f_{\bm{\theta}_{t}}, \mc{D}_t)+\sum_{k=1}^{t-1}\alpha_k\ell(f_\theta,f_{\bm{\theta}_{k}}, \mc{M}_k)
  \label{eq:er}
\end{equation}
where $\alpha$ is the weight for each task-specific loss.
This kind of method relies on the effective weighted summation of all gradients, but ignoring the ensembled gradients may conflict with some old tasks' gradients.
In addition, the magnitude of new tasks' gradients exceeds that of old tasks in most cases, which leads to the summed gradient having a large bias toward the new tasks. 
Accordingly, rehearsal may make old tasks drifting and entangled with each other, causing domain chaos (see Fig.~\ref{fig:drift}(b)).

The above two kinds of rehearsal-based methods in Eqs.~\eqref{eq:eps} and \eqref{eq:er} ignore the origin of catastrophic forgetting in rehearsal-based OCL, and suffer from feature entanglement between old and new tasks.
In the following section, we will propose a new paradigm of rehearsal-based OCL, named drift-reducing rehearsal, which analyzes and mitigates the continual domain drift with two proposed inequalities.

\section{Drift-Reducing Reherasal}

\subsection{Overview: Continual Domain Drift}

\label{sec:overview}


In rehearsal-based OCL, suppose all tasks are independent and identically distributed (I.I.D.). 
That is, a domain drift for task $i$ at task $t>1$ can be denoted as the feature drift from $f_i^{t-1}$ to $f_i^{t}$.
If $i=t$, $f_i=f_{\bm{\theta}}(\mc{D}_i)$ represents the feature on the current data for the new task.
If $i<t$, $f_i=f_{\bm{\theta}}(\mc{M}_i)$ represents the feature on the memory buffer for the old tasks.
To reduce the drift, we transform the problem to two minimizing distance constraints.
First, for the old task $i\in[1,t)$, let $d(\cdot)$ denote the distance between two domains, and the problem needs to satisfy the following stability inequality:
\begin{equation}
  d(f^t_i,f^{i}_i) \le d(f^{t-1}_i,f^{i}_i).
	\label{eq:dist1}
\end{equation}
Eq.~\eqref{eq:dist1} means that the distance from the new location to the original location in the feature space for an old task should not get larger.
Second, for $i,j\in[1,t]$ and $i\ne j$, the domain distance across tasks should satisfy the following inequality:
\begin{equation}
		d(f^t_i,f^t_j) \ge 
    \left\{
      \begin{aligned}
        d(f^{t-1}_i,f^{t-1}_j) \ge m, \quad \text{Old and old}.\\
        m, \quad \text{New and old}
      \end{aligned}
    \right.
	\label{eq:dist2}
\end{equation}
where $m$ means a pre-defined margin.
For the $t$ tasks w.r.t. datasets $\{\mc{D}_1,\cdots,\mc{D}_t\}$, the Inequality~\eqref{eq:dist2} means that the domain distance between any two tasks should not be smaller than the model trained on the last task.



However, the distance can be evaluated only if we further store every previous model when training with new tasks, which is costly in OCL.
Thus, we further transform the two above inequalities into two goals.
First, we should store more representative samples in rehearsal to reduce the influence of unpredictable drift from out-of-distribution samples.
Second, we should enlarge the discrepancy between different tasks and different classes.

\emph{To achieve this, corresponding to the above two goals, we design to reduce the continual domain drift in OCL via constructing centroids}.
Centroid refers to the center point of a cluster for each class or task. 
In OCL, where the model is updated continuously as new data becomes available, the centroid of a cluster may change over time as new data points are added.
As new data points arrive, the centroid of a cluster may be updated to reflect the changes in the distribution of the data. This allows the model to adapt to evolving patterns in the data over time.
For the first goal, it is feasible to leverage centroids to select more representative data points in the process of rehearsal, thus approaching the original data distribution.
For the second goal, one can further constrain the drift of each task based on the constructed centroids.


Accordingly, we propose a novel Drift-Reducing Rehearsal (DRR) method to disentangle continual domain drift and reduce the negative transfer.
Our DRR reduces the continual domain drift in OCL via two key components:

(1) \textit{Centroid-based Online Selection (COS)}: 
As the distribution shifts from the entire training dataset to a small memory buffer, the stored memory buffer has poor generalization as the whole training set.
Thus, we design to store more representative data according to centroids in an online fashion.

{(2) \textit{Contrastive Margin Loss (CML)}:
We propose to connect each isolation task and set two-level angular margins within the contrastive loss framework. This is achieved through the use of constructed centroids, which serves to enhance intra-class/task compactness and inter-class/task discrepancy.}

{We illustrate the framework of the proposed DRR method in Fig.~\ref{fig:method}.}
In OCL, when a new task comes with a new mini-batch, DRR builds or updates the centroids and the stored data in memory via COS. For the stored data, we train them with the new tasks and connect their logits by the proposed cross-task CML.
In the following subsections, we will illustrate the two components in detail.


\subsection{{Centroid-based Online Selection}}

To encourage the selection, we propose a novel centroid-based online selection method, called COS, to store more representative samples in the streaming training data of OCL.
Building centroids in OCL is equivalent to adaptive data clustering, but relies on a pre-defined centroid distance threshold $\epsilon$.
Many previous methods construct centroids to represent data in previous studies~\cite{ayub2019centroid,ayub2020cognitively}.
To align the construction with the OCL setting, three critical factors must be addressed:
1) \emph{Continual task evolution}: The task evolves continuously, necessitating the rapid generation of appropriate centroids.
2) \emph{Adaptation to domain drift}: The domain drifts for old tasks, necessitating ongoing updates to the centroids to align with the evolving distribution.
3) \emph{Limited memory budget}: With a fixed memory buffer budget, the centroids must be updated judiciously as the number of tasks increases. 
Motivated by these considerations, our COS strategy, designed for a data stream, encompasses three main steps.

\textit{(1) Build new centroids}:
For any new tasks, let $\mb{c}_{i}$ be the $i$-th centroid. If there exists no centroid that satisfies the distance constraint to a new-input data point, \ie, 
\begin{equation}
  \mathop{\min}_{i} \left\| f_{\bm{\theta}}\left(x\right) - \mb{c}_{i} \right\|>\epsilon
\end{equation}
we set a new centroid according to this data point as
\begin{equation}
  \mb{c}_\text{new} = f_{\bm{\theta}}\left(x\right).
\end{equation}
The creation of a new centroid is simple and similar to many traditional centroid methods.

\textit{(2) Update existing centroids}:
Otherwise, for the same new task, if we have 
\begin{equation}
  \mathop{\min}_{i} \left\| f_{\bm{\theta}}\left(x\right) - \mb{c}_{i} \right\|\le\epsilon
\end{equation}
which means that this data point belongs to an existing centroid with the smallest distance.
We then update the target centroid of the new task by 
\begin{equation}
  \mb{c}_{i^*}=\frac{N_{i^*}^{C} \times \mb{c}_{i^*}+f_{\bm{\theta}}\left(x\right)}{N_{i^*}^{C}+1}
\end{equation}
where 
\begin{equation}
  i^* = \mathop{\arg\min}_i \left\|f_{\bm{\theta}}\left(x\right) - \mb{c}_{i} \right\|.
\end{equation}
Note that the centroid construction of a new task is greedy, which means that we need to generate enough centroids as quickly as possible.
This can be achieved by carefully selecting the threshold $\epsilon$.

\begin{figure}[t]
	\centering
	\includegraphics[width=\linewidth]{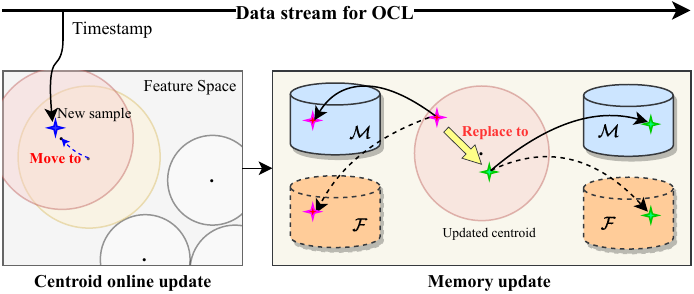}
  \vspace{-15px}
	\caption{
		 Centroid-based online selection. For each data point on a data stream, we first compare it with existing centroids if it satisfies the threshold limit. Left: If the data point is captured by a centroid with the smallest allowed distance, the target centroid will be updated.
     Right: Then, we reevaluate the distance to each centroid, and replace the farthest point with the new-captured sample in the memory buffer.
	}
	\label{fig:sample}
\end{figure}

\textit{(3) Update memory with reallocation}:
For any old tasks, given the fixed-size memory buffer $\mc{M}$, we divide it evenly for each class.
For class $C$, its corresponding memory $\mc{M}^C$ can be sampled from all related centroids. 
A sample from the $i$-th centroid will be selected under the Bernoulli distribution $B(1, p(i))$, and the probability is computed by
\begin{equation}
 p(i)= \frac{{N^C_i}}{{\sum_j{N^C_j}}}
 \label{eq:pi}
\end{equation} 
where the value $N_i^C$ means the number of seen samples by the $i$-th centroid.
The whole selection pipeline can be found in Algorithm~\ref{alg:centroid} and Fig.~\ref{fig:sample}.
The sampled data is stored in memory $\mc{M}^C=\{(x^C_{1},y^C_{1}),\cdots,(x^C_{|M^C|},y^C_{|M^C|})\}\subset\mc{D}_C$.

To enhance the ability to reduce continual domain drift, we also leverage a simple yet effective optional strategy for old tasks, named Centroid Distilllation Loss (CDL) based on the above centroids.
CDL is designed to reduce the continual domain drift along with the continual learning process.
First, apart from the raw data, we also store the corresponding feature representations from its current model, denoted as $\mc{F}^C=\{\mb{f}^C_{1},\cdots,\mb{f}^C_{|M^C|}\}$.
The CDL is defined as
\begin{equation}
	{\ell}_{\text{D}}(f_\theta(x_i^C), \mb{f}_i^C)=\left\|f_\theta(x_i^C)- \mb{f}_i^C\right\|^2.
	\label{eq:lossf}
\end{equation}
The CDL can be in many formats, and we choose the Mean Square Error (MSE). 
By training with Eq.~\eqref{eq:lossf} in each step, we can ease the continual domain drift effectively.

In general, the proposed COS strategy can select representatives in a data stream without storing the whole dataset.
The CDL builds extra memory buffers to save the feature representation for each stored sample.
We find that the extra memory buffers are very small in size compared to the whole training set.
In our implementation, we save the representation from the last fully-connected layer before the classifier, which is a vector with lengths from 256 to 2048 (for different backbones).
This means that the memory cost is smaller than saving the raw data (\eg, $32\times 32 \times 3$ for CIFAR).

\begin{algorithm}[tb]
	\caption{{Centroid-based Online Selection (COS)}}
	\label{alg:centroid}
	\begin{algorithmic}[1]
		\REQUIRE {\textsc{COS}} (Memory buffer $\mc{M}$, Feature memory buffer $\mc{F}$, Input data $(x, y)$, Model $f$)\\\textbf{Known items}: Seen class set $\mc{C}$,  Threshold $\epsilon$, Centroid count $\{N_{1:|\mc{C}^y|}^y\}$, Buffer size $m$
    \STATE $i^* = \mathop{\arg\min}_{i} \left\| f_{\bm{\theta}}\left(x\right) - \mb{c}_{i} \right\|$
    \STATE $d = \left\| f_{\bm{\theta}}\left(x\right) - \mb{c}_{i^*} \right\|$
    \IF{$d>\epsilon$}
    \STATE $\mc{C}_y\leftarrow\mb{c}_\text{new} = f_{\bm{\theta}}\left(x\right)$$\hfill$\texttt{\color{blue}\#Create new centroid}
    \STATE $N_{|\mc{C}^y|+1}^{y} = 1$
    \ELSE
    \STATE $\mb{c}_{i^*}=\frac{N_{i^*}^{y} \times \mb{c}_{i^*}+f_{\bm{\theta}}\left(x\right)}{N_{i^*}^{y}+1}$ $\hfill$\texttt{\color{blue}\#Update centroid}
    \STATE $N_{i^*}^{y} = N_{i^*}^{y}+1$
    \ENDIF
    \STATE \texttt{\color{blue}\#Update memory with sampling}
    \STATE $s\sim B(1, p(i))$ $\hfill$\texttt{\color{blue}\#Sample with probability~\eqref{eq:pi}}
    \IF{$s=1$} 
    \STATE \texttt{\color{blue}\#Add new samples}
		\STATE $\mc{M}^y\leftarrow\mc{M}^y\cup \{({x},y)\}$
		\STATE $\mc{F}^y\leftarrow\mc{F}^y\cup \{f(\mb{x})\}$ 
    \ENDIF
    \IF{$\left|\mc{M}^y\right|> \cfrac{m}{|\mc{C}|}$} 
    \STATE \texttt{\color{blue}\#Remove extra samples}
    \STATE $x'_m,y'_m = \mathop{\arg\max}_{(x',y')\in\mc{M}^y_{i^*}} \left\| f_{\bm{\theta}}\left(x'\right) - \mb{c}_{i^*} \right\|$
    \STATE Remove ($x'_m,y'_m$) and $f_{\bm{\theta}}\left(x'_m\right)$ from $\mc{M}$ and $\mc{F}$
    \ENDIF
		\STATE {\bfseries Return} $\mc{M}$, $\mc{F}$
		
	\end{algorithmic}
\end{algorithm}

\subsection{Contrastive Margin Loss}

By the aforementioned COS, we store more representative samples in the process of training CL model. 
However, only representative memory is not enough to solve continual domain drift, because the number of samples is too small than the training set.
It is better to further constrain the latent continual domain drift in the feature space.
In contrast to traditional proxy-based learning with softmax classifier, using contrastive-based loss and nearest class mean (NCM) classifier witness better generalization~\cite{cha2021co2l}.
However, the contrastive loss is subjected to the ``bias'' issue caused by class imbalance, tending to classify most samples of old classes into new categories~\cite{lin2023pcr}. 
We claim that this kind of bias makes it difficult to reduce continual domain drift in two aspects.
First, the traditional contrastive loss cannot observe the drift of the same class from past to present.
Second, task imbalance between different tasks exists in rehearsal-based CL, and traditional contrastive loss cannot avoid drifting from this kind of task imbalance.
In this paper, we propose to extend the traditional contrastive loss for each task to disentangle continual domain drift based on the centroid extracted from COS.

Given data point $(x,y)$ with its feature $\mb{z}=f_{\bm{\theta}}(x)$, we have the traditional contrastive loss for a mini-batch
\begin{equation}
	\ell_k(x)=\sum_{i \in \mc{B} }\frac{-1}{|\mc{P}(i)|} \sum_{p \in \mc{P}(i) } \log \frac{e^{\mb{z}_i^\top \mb{z}_p/\tau}}{\sum_{j \in \mc{B} \text{\textbackslash} \{i\}}e^{\mb{z}_i^\top \mb{z}_j/\tau}}
	\label{eq:scl}
\end{equation}
where $\mc{B}$ denotes a mini batch and $\mc{B} \text{\textbackslash} \{i\}$ means the batch excludes the sample $i$.
$\mc{P}(i) = \{ p \in A(i): y_p = y_i \}$ is the set of all positives (i.e., samples with the same labels as sample $i$ ) in batch excluding sample $i$.
The contrastive loss is built across tasks.
That is, each classifier leverages the features from other tasks to improve the perception between tasks.
The cross-task contrastive loss helps to equally treat new tasks and old tasks. 

Nonetheless, the contrastive softmax classifier is not effective because it ignores the intra-class compactness and the inter-class discrepancy, which is important for the classification in drift space.
In recent years, some methods such as~\cite{deng2018arcface,liu2016large-margin} use large margins in the classifier to improve the discriminative capability.
Inspired by the well-proven Arcface~\cite{deng2018arcface}, we also seek to add large margins to the angle between contrastive features.
Specifically, Arcface deletes the bias of the last full connected (FC) layer and transforms $\mb{W}_{j}^{T} \mb{x}=\left\|\mb{W}_{j}\right\|\left\|\mb{x}\right\| \cos \beta_{j}$, where $\beta_{j}$ denotes the angle between the weight $\mb{W}_{j}$ and the feature vector $\mb{x}$.
Then an angular margin $m$ is placed between class $y$ and other classes
\begin{equation}
	\ell(x)=-\log \frac{e^{s\cdot\cos \left(\beta_{y}+m\right)}}{e^{s\cdot\cos \left(\beta_{y}+m\right)}+\sum_{j=1, j \neq y}^{n} e^{s\cdot \cos \beta_{j}}}
	\label{eq:arcface}
\end{equation}
where the magnitude of individual weight $\|\mb{W}_j\|$ is fixed to $1$ and the feature $\mb{x}$ is rescaled to $s$ by $l_2$-norm.
The normalization makes the predictions magnitude-irrelevant but improves the intra-class compactness and inter-class discrepancy.

However, Arcface cannot be directly applied to our cross-task contrastive loss, because the method places large margins only on a single task in softmax function.
We prefer to reduce the possible interference from new tasks to old tasks, \ie, catastrophic forgetting, and the negative transfer from old tasks to new tasks.
In this paper, we propose two levels of margins for each classifier \ie, \textit{class level} and \textit{task level} margins in contrastive loss.
Based on Eq.~\eqref{eq:arcface}, we propose our \emph{Contrastive Margin Loss} (CML).

First, for two samples $x_1$ and $x_2$, we have their feature relations between $\mb{z}_1$ and $\mb{z}_2$ computed by
\begin{equation}
	\mb{z}_1^\top \mb{z}_2 = \| \mb{z}_1 \|  \| \mb{z}_2 \| \cos(\beta_{1,2}).
\end{equation}
The function means that the contrastive relations between two samples can be projected to an angle space.
Inspired by Arcface, we place the two margins in this angle space.
Specifically, for the task $k\in[1,t]$, we have CML computed as follows:
\begin{equation}
	\ell_{k}(x)=\sum_{y \in I }\frac{-1}{|P(y)|} \sum_{p \in P(y) } \log \frac{e^{s\cdot\cos \left((\beta^k_{y,p}+m^\text{c})+m^\text{t}\right)}}{\sigma_\text{c}+\sigma_\text{t}+\sigma_\text{s}}
	\label{eq:dmsl}
\end{equation}
where 
\begin{equation}
		\sigma_\text{c} =  {e^{s\cdot\cos \left((\beta_{{y,p}}^k+m^\text{c})+m^\text{t}\right)}} ,~~\sigma_\text{t} =  {\sum_{j=1, j \neq y}^{C_k} e^{s\cdot \cos (\beta_{y,j}^k+m^\text{t})}}
  \end{equation}
denote the class-level and task-level similarities in Eq.~\eqref{eq:dmsl}, respectively.
Moreover, others comparisons are denoted as 
\begin{equation}
    \sigma_\text{s} =  {\sum_{i=1, i \neq k}^{t}\sum_{j=1}^{C_i} e^{s\cdot \cos \beta_{y,j}^i}}.
\end{equation}

\begin{algorithm}[t]
	\caption{Drift-Reducing Rehearsal (DRR)}
	\label{alg:method}
	\begin{algorithmic}[1]
		\REQUIRE \textsc{Train}($f_\theta$, $f_{\theta_{1:T}}$, $\{\mc{D}_1^{\text{trn}},\cdots,\mc{D}_T^{\text{trn}}\}$)
		
		\STATE $\mc{M},\mc{F}\leftarrow\{\},\{\}$$\hfill$\texttt{\color{blue}\#Memory initialization}
		\FOR{$t=1$ {\bfseries to} $T$}
		\FOR{mini-batch $\mc{B}\in\mc{D}_t^{\text{trn}}$}
		\STATE $g,g_t\leftarrow\nabla_\theta\ell_t(f_\theta, f_{\theta_t}, \mc{B})$$\hfill$\texttt{\color{blue}\#CML for new task}
		\IF{$t=1$}
		\STATE $\tilde{g}\leftarrow g$ 
		\ELSE
    \STATE \texttt{\color{blue}\#CML for old tasks}
		\STATE $g^\text{ref},g_{1:t-1}\leftarrow\nabla_\theta\ell_{1:t-1}(f_\theta, f_{\theta_{1:t-1}}, \mc{M})$
    \STATE \texttt{\color{blue}\#CDL for old tasks}
		\STATE $g^\text{ref}\leftarrow g^\text{ref} + \nabla_\theta{\ell}_\text{D}(f_\theta, \mc{F}_\text{ref})$ 
		\STATE $\tilde{g}\leftarrow g+g^{\text{ref}}$
		\ENDIF		
    \STATE \texttt{\color{blue}\#Update shared parameter}
		\STATE $\theta\leftarrow\theta-\text{StepSize}\cdot\tilde{g}$
    \STATE \texttt{\color{blue}\#Update task-specific parameters}
		\STATE $\theta_{1:t}\leftarrow\theta_{1:t}-\text{StepSize}\cdot g_{1:t}$
		\ENDFOR
		\STATE $\mc{M},\mc{F}\leftarrow\text{\textsc{COS}}(\mc{M},\mc{F},\mc{D}_t^{\text{trn}},f_\theta)$
		\ENDFOR
	\end{algorithmic}
\end{algorithm}

Concretely, $m^\text{c}$ denotes the class-level margin for the ground-truth class.
$m^\text{t}$ represents the task level margin for the current incremental learning task.
In Eq.~\eqref{eq:dmsl}, we add class-level margin $m^\text{c}$ and task-level margin $m^\text{t}$ to the angular $\beta$.
$m^\text{c}$ is similar to $m$ in Eq.~\eqref{eq:arcface}, which controls the intra-task class compactness and discrepancy~\cite{deng2018arcface}.
{$m^\text{t}$ controls the task compactness and discrepancy, which ensures the knowledge of each task does not mix up with others.}

CML offers two distinct advantages.
First, it enhances task boundary clarity, effectively segregating old and new tasks, thereby demonstrating the efficacy of $m_\text{t}$. 
Second, CML alleviates domain overlap resulting from continual domain drift by compelling tasks to diverge in angular space. 
The implementation of two-tier angular margins for old and new tasks enhances both intra-class/task compactness and inter-class/task discrepancy.


\subsection{Total Algorithm}

We unites the memory of all old tasks for efficiency.
Let $\mc{M}=\cup_{k<t}\mc{M}_k$ and $\mc{F}=\cup_{k<t}\mc{F}_k$ be the united data and representation memory for old tasks.
Then, we train the model with an updated objective as
\begin{equation}
	\ell = \underbrace{\ell_t(f_\theta, f_{\theta_t}, \mc{D}_t)}_{\text{CML: new task}}+
	\underbrace{\ell_{1:t-1}(f_\theta, f_{\theta_{1:t-1}}, \mc{M})}_{\text{CML: rehearsal on old tasks}}+\underbrace{{\ell}_{\text{D}}(f_\theta, \mc{F})}_{\text{CDL~(optional)}}.
	\label{eq:finalloss}
\end{equation}

{We show the detailed OCL training in Algorithm~\ref{alg:method}, where we have each mini-batch pass only once.
For each mini-batch, we first compute the CML for the new task, note that the task-level margin is not active if $t=1$.
Then, for $t>1$, we further compute the CML and CDL for old tasks to reduce catastrophic forgetting.
The three losses are used to compute the gradient of shared and task-specific parameters, and we use the ensemble gradient to update the model.
After the model update, the storage of memory feature in \textsc{COS} becomes the anchor of old task in current task training.  }

\section{Experiments}

\subsection{Datasets}


We evaluate our DRR method on four popular CL datasets.

\noindent
\textbf{Permuted MNIST} \cite{kirkpatrick2017overcoming} is a variant of the MNIST dataset where the pixel locations are randomly permuted to create new sequences of images. This variation challenges models to adapt to different pixel permutations, thus testing their robustness and plasticity to input variations. It is commonly used to evaluate model performance in the face of distribution shifts, particularly in tasks such as transfer learning and continual learning.
  We set 20 fixed permutations on pixels of all images to stand for 20 different tasks.
  Each task has a fixed random permutation of the input pixels which is applied to all the images of that task.
  \\
  \textbf{Split CIFAR} \cite{zenke2017continual} is a variant of the CIFAR-100 dataset where the classes are divided into different subsets, each subset used for a different task. 
   This dataset is often employed to assess model performance in CL, simulating real-world scenarios where the data distribution changes over time.   
  We randomly and evenly splits CIFAR-100 dataset into 20 tasks, where each task has 5 classes.
  \\
  \textbf{Split CUB.} \cite{AGEM} is a variant of the CUB-200-2011 bird dataset where the bird categories are divided into different subsets like Split CIFAR.
  We splits the CUB dataset~\cite{wah2011caltech} randomly and evenly into 20 tasks, where each task has 10 classes.
  \\
  \textbf{Split AWA} \cite{AGEM} is a variant of the Animals with Attributes (AWA) dataset where animal categories are divided into different subsets, each subset is used for a different task.  This dataset consists of 20 subsets of the AWA dataset~\cite{lampert2009learning}, where each subset has 5 classes with replacement from a total of 50 classes. 
  Note that the same class can appear in different subsets. As in \cite{AGEM}, in order to guarantee that each training example only appears once in the learning process, based on the occurrences in different subsets the training data of each class is split into disjoint sets.


\subsection{Evaluation Metrics}

In our paper, we evaluate OCL methods using three metrics.

\noindent
\textbf{Average Accuracy.}~($A_t\in[0,1]$)
  For task $t\in\{1,\cdots,T\}$, the average accuracy is computed as follows.
  \begin{equation}
    A_{t}=\frac{1}{t} \sum_{j=1}^{t} a_{B_{t}, j}
  \end{equation}
  where $a_{B_{t}, j}$ is the mean accuracy of task $j$ on $B_{t}$ mini-batches.
  In particular, $A_T$ is the final average accuracy on all tasks.
  \\
  \textbf{Forgetting Measure}~\cite{chaudhry2018riemannian}.~($F_t\in[-1,1]$) 
  For task $t\in\{1,\cdots,T\}$, the average forgetting is defined as follows.
  \begin{equation}
    F_{t}=\frac{1}{t-1} \sum_{j=1}^{t-1} f_{j}^{t}
  \end{equation}
  where $f_{j}^{t}$ is the forgetting on task $j$ after the model finishing the training of task $t$ and is computed as
  \begin{equation}
    f_{j}^{t}=\max _{l \in\{1, \cdots, k-1\}} a_{l, B_{l}, j}-a_{t, B_{k}, j}.
  \end{equation}
  This metric is also known as backward transfer (BWT).
  \\
  \textbf{Long-Term Remembering (LTR)}~\cite{guo2019learning}.~($\text{LTR}\ge0$)
  LTR evaluates the accuracy drop on each task relative to the accuracy just right after the task has been learned. The final LTR is defined as
  \begin{equation}
    \mathrm{LTR}=-\frac{1}{T-1} \sum_{j=1}^{T-1}(T-j) \min \left\{0, a_{T, B_{T}, j}-a_{j, B_{j}, j}\right\}.
  \end{equation}
  LTR quantifies the accuracy drop on task $D_j$ relative to $a_{j, B_{j}, j}$. 

\begin{table}[t]
  \centering
  \caption{{Comparison with SOTAs on Permuted MNIST.}}
  \label{tab:sota_mnist}
  \resizebox{\linewidth}{!}{
    \begin{tabular}{lcccc}
      \toprule
      \textbf{Method}  & {$A_{{T}}(\%)$} & {$F_{{T}}$} & {$LTR$} \\
      \midrule
      JOINT (UB)   	& $95.30$ & - & -  \\
      VAN (LB)    	& $47.55\pm2.37$ & $0.52\pm0.026$  & $5.375\pm0.194$  \\
      \midrule
      EWC 	& $68.68\pm0.98$ & $0.28\pm0.010$  & $3.292\pm0.135$  \\
      MAS    	& $70.30\pm1.67$ & $0.26\pm0.018$  & - \\
      RWalk   & $85.60\pm0.71$ & $0.08\pm0.007$  & -  \\
      GEM     & $89.50\pm0.48$ & $0.06\pm0.004$ & - \\
      A-GEM   & $89.32\pm0.46$ & $0.07\pm0.004$  & $0.716\pm0.048$  \\
      ER   	& $90.47\pm0.14$ & $0.03\pm0.001$  & $0.367\pm0.013$  \\
      MEGA-I   	& $91.10\pm0.08$ & $0.05\pm0.001$  & -  \\
      MEGA-II   & $91.21\pm0.10$ & $0.05\pm0.001$ & - \\
      $\epsilon$-SOFT-GEM   	& $91.30\pm0.11$ & $0.05\pm0.001$   & -  \\
      DER   	&$92.03\pm0.19$ &$0.04\pm0.001$  &$ 0.402\pm0.012$ \\
      SCR   	& $ 91.74\pm0.63$ & $ 0.05\pm0.004$ &$ 0.492\pm0.041$  \\
      MDMT-R   	& $94.33\pm0.04$ & ${0.02}\pm0.000$  & $0.247\pm0.009$  \\
      \midrule
      DRR (Ours)   &$ \textbf{94.43}\pm0.17$ &$\textbf{0.02}\pm0.001$ &$ \textbf{0.243}\pm0.001$ \\
      \bottomrule
    \end{tabular}}
\end{table}

\begin{table}[t]
  \centering
  \caption{Comparison with state-of-the-arts on Split CIFAR.}
  \label{tab:sota_cifar}
  \resizebox{\linewidth}{!}{
    \begin{tabular}{lcccc}
      \toprule
      \textbf{Method}  & {$A_{{T}}(\%)$} & {$F_{{T}}$} & {$LTR$} \\
      \midrule
      JOINT (UB)   	& $68.30$ & - & -  \\
      VAN (LB)    	& $40.44\pm1.02$ & $0.27\pm0.006$  & $2.613\pm0.174$ \\
      \midrule
      EWC 	& $42.67\pm4.24$ & $0.26\pm0.039$  & $2.493\pm0.427$  \\
      MAS    	& $42.35\pm3.52$ & $0.26\pm0.030$  & - \\
      RWalk     & $42.11\pm3.69$ & $0.27\pm0.032$  & - \\
      MER       & $37.27\pm1.68$ & $0.03\pm0.030$  & - \\
      GEM       & $61.20\pm0.78$ & $0.06\pm0.007$  & - \\
      A-GEM     & $61.28\pm1.88$ & $0.09\pm0.018$  & $0.643\pm0.124$  \\
      ER   	& $63.97\pm1.30$ & $0.06\pm0.006$  & $0.451\pm0.333$ \\
      MEGA-I   	& $66.10\pm1.67$ & $0.05\pm0.014$  & -   \\
      MEGA-II   & $ 66.12\pm1.94$ & $0.06\pm0.015$  & -  \\
      $\epsilon$-SOFT-GEM   	& $63.90\pm1.53$ & $0.06\pm0.014$  & -  \\
      DER   	& $68.49\pm1.45$ & $0.06\pm0.009$  & $0.371\pm0.087$  \\
      SCR   	& $67.99\pm1.89 $ & $ 0.05\pm0.004$ & $ 0.258\pm0.024$  \\
      ASER   	& $65.53\pm1.89 $ & $ 0.07\pm0.007$  & $ 0.544\pm0.133$  \\
      MDMT-R   	& $69.20\pm1.60$ & $0.04\pm0.010$  & $0.283\pm0.099$  \\
      \midrule
      DRR (Ours)   & $\textbf{72.47}\pm2.34$ & $\textbf{0.03}\pm0.011$  & $\textbf{0.127}\pm0.093$  \\
      \bottomrule
    \end{tabular}}
\end{table}


\subsection{Implementation Details}

\noindent
\textbf{Network architectures.}
Following previous studies~\cite{lopez2017gradient,AGEM,guo2019learning,lyu2021multi}, we use a FC network with two hidden layers for Permuted MNIST, where each layer has 256 units with the ReLU activation. 
For Split CIFAR, we use a reduced ResNet18 as in~\cite{he2016deep}. 
For Split CUB and Split AWA, we use a standard ResNet18.

\noindent
\textbf{Seed initialization.} 
We set 5 fixed seeds from 1,234 to 1,238 for all compared methods, and the show the average accuracy and the standard deviation.


\noindent
\textbf{Hyper-Parameters selection.}
We report the hyper-parameters used in our experiments.
Following A-GEM~\cite{AGEM}, the learning rates for each dataset are $0.1$ for Permuted MNIST, $0.03$ for Split CIFAR and CUB, and $0.01$ for Split AWA.
The selected centroid thresholds $\epsilon$ are $6$ for Permuted MNIST, $8$ for Split CIFAR and AWA, and $7$ Split CUB.
The hyperparameter $m_\text{t}$ is set to 0.1 for Permuted MNIST and Split CIFAR, 0.2 for Split AWA, and 0.4 for Split CUB.
We set $m_\text{c}$ to 0.01 for Permuted MNIST and Split CIFAR, 0.02 for Split AWA, and 0.05 for Split CUB.
we set $s$ to 20 for Split CUB, 24 to Split CIFAR, and 32 for Permuted MNIST and Split AWA.

\begin{table}[t]
  \centering
  \caption{Comparison with SOTAs on Split CUB.}
  \label{tab:sota_cub}
  \resizebox{\linewidth}{!}{
    \begin{tabular}{lcccc}
      \toprule
      \textbf{Method}  & {$A_{{T}}(\%)$} & {$F_{{T}}$}  & {$LTR$} \\
      \midrule
      JOINT (UB)   	& $65.60$ & -  & -  \\
      VAN (LB)   	& $53.89\pm2.00$ & $0.13\pm0.020$  & $0.976\pm0.215$ \\
      \midrule
      EWC 	& $53.56\pm1.67$ & $0.14\pm0.024$ &  $1.021\pm0.210$ \\
      MAS    	& $54.12\pm1.72$ & $0.13\pm0.013$  & - \\
      RWalk     & $54.11\pm1.71$ & $0.13\pm0.013$  & - \\
      SI        & $55.04\pm3.05$ & $0.12\pm0.026$  & - \\
      A-GEM     & $61.82\pm3.72$ & $0.08\pm0.021$  & $0.456\pm0.174$   \\
      ER   	& $73.63\pm0.52$ & ${0.01}\pm0.005$  & $\textbf{0.001}\pm0.001$ \\
      MEGA-I   	& $79.67\pm2.15$ & $0.01\pm0.019$  & -  \\
      MEGA-II  	& $80.58\pm1.94$ & $0.01\pm0.017$  & -  \\
      $\epsilon$-SOFT-GEM   	& $75.60\pm2.00$ & $0.03\pm0.009$  & -  \\
      DER   	& $76.56\pm2.48$ & $0.01\pm0.015$  & $0.025\pm0.018$  \\
      SCR   	& $ 81.43\pm1.97$ & $0.01\pm0.007$  & $ 0.007\pm0.009$  \\
      ASER   	& $ 75.58\pm3.72 $ & $ 0.02\pm0.010$  & $0.037\pm0.029$  \\
      MDMT-R   	& $84.27\pm1.63$ & $0.01\pm0.015$  & $0.017\pm0.014$  \\
      \midrule
      DRR (Ours) & $\textbf{85.71}\pm1.26$ & $\textbf{0.01}\pm0.005$  & $0.007\pm0.007$  \\
      \bottomrule
    \end{tabular}}
\end{table}

\begin{table}[t]
  \centering
  \caption{Comparison with state-of-the-arts on Split AWA.}
  \label{tab:sota_awa}
  \resizebox{\linewidth}{!}{
    \begin{tabular}{lcccc}
      \toprule
      \textbf{Method}  & {$A_{{T}}(\%)$} & {$F_{{T}}$}  & {$LTR$} \\
      \midrule
      JOINT (UB)   	& $64.80$ & -  & -  \\
      VAN (LB)   	& $30.35\pm2.81$ & $0.04\pm0.013$  & $0.202\pm0.090$  \\\hline
      EWC 	& $33.43\pm3.07$ & $0.08\pm0.021$  & $0.675\pm0.214$ \\
      MAS    	& $33.83\pm2.99$ & $0.08\pm0.022$  & - \\
      RWalk   & $33.63\pm2.64$ & $0.08\pm0.023$  & -  \\
      SI     & $33.86\pm2.77$ & $0.08\pm0.022$  & - \\
      A-GEM   & $44.95\pm2.97$ & $0.05\pm0.014$  & $0.178\pm0.082$   \\
      ER   	& $54.27\pm4.05$ & ${0.02}\pm0.030$  & $0.014\pm0.015$  \\
      MEGA-I   	& $54.82\pm4.97$ & $0.04\pm0.034$  & -  \\
      MEGA-II   	& $54.28\pm4.84$ & $0.05\pm0.040$  & -  \\
      $\epsilon$-SOFT-GEM   	& $55.30\pm3.57$ & $0.01\pm0.028$  & -  \\
      DER   	& $50.70\pm4.91$ &$0.04\pm0.040$  & $0.063\pm0.094$  \\
      SCR   	& $54.35\pm2.68$ & $ 0.02\pm0.012$  & $ 0.022\pm0.010$  \\
      ASER  	& $46.72\pm3.20$ & $ 0.05\pm0.006$  & $ 0.171\pm0.021$  \\
      MDMT-R   	& $61.56\pm3.36$ & $0.02\pm0.027$  & $\textbf{0.002}\pm0.002$  \\
      \midrule
      DRR (Ours) 	& $\textbf{62.86}\pm4.04$ & $\textbf{0.01}\pm0.005$ & $0.009\pm0.011$  \\
      \bottomrule
    \end{tabular}}
\end{table}

\begin{table}[ht]
  \centering
  \caption{Ablation study of margins on Split CIFAR.}
  \label{tab:abl}
  \resizebox{\linewidth}{!}{
    \begin{tabular}{cccccc}
      \toprule
      $m_\text{t}$ & $m_\text{c}$ & CDL & {$A_{{T}}(\%)$} & {$F_{{T}}$}  & {$LTR$}  \\
      \midrule
      - & - &  & $66.23\pm1.35$ & $0.046\pm0.006$  & $0.314\pm0.052$  \\\midrule
      - & - & $\checkmark$ & $67.84\pm2.54$ & $0.041\pm0.017$  & ${0.287}\pm0.082$  \\\midrule
      0.0 & 0.0 &  & $71.83\pm1.76$ & $0.036\pm0.005$  & $0.206\pm0.056$ \\
      0.1 & 0.0 &  & $72.01\pm2.65$ & $\mb{0.034}\pm0.005$  & $\mb{0.187}\pm0.016$ \\
      0.0 & 0.01 &  & $69.99\pm1.71$ & $0.034\pm0.016$  & $0.233\pm0.128$ \\
      0.1 & 0.01 &  & $72.01\pm2.13$ & $0.031\pm0.015$  & $0.161\pm0.078$ \\
      0.4 & 0.01 &  & $71.55\pm1.67$ & $0.032\pm0.008$  & $0.186\pm0.065$ \\
      0.4 & 0.05 &  & $71.81\pm2.82$ & $0.035\pm0.007$  & $0.214\pm0.053$ \\
      0.4 & 0.1 &  & $71.56\pm1.67$ & $0.043\pm0.008$  & $0.216\pm0.066$ \\
      \midrule
      0.1 & 0.01 & $\checkmark$ & $\mb{72.47} \pm2.34$ & $0.027 \pm0.011$  & $0.127 \pm0.093$ \\
      \bottomrule
    \end{tabular}}
\end{table}

\begin{table}[ht]
	\centering
	\caption{Ablation study of centroid threshold $D$ on Split CIFAR.}
	\label{tab:abl2}
	\resizebox{.8\linewidth}{!}{
		\begin{tabular}{cccc}
			\toprule
			$D$ & {$A_{{T}}(\%)$} & {$F_{{T}}$}  & {$LTR$}  \\
			\midrule
			- & $70.62\pm1.60$ & $0.042\pm0.010$  & $0.283\pm0.099$  \\
			\midrule
			6.0 & $71.72\pm2.34$ & $0.034\pm0.010$  & $0.172\pm0.092$ \\
			6.5 & $72.02\pm2.37$ & $0.032\pm0.010$  & $0.178\pm0.059$ \\
			7.0 & $72.12\pm2.52$ & $0.039\pm0.008$  & $0.165\pm0.061$ \\
			7.5 & $72.28\pm2.11$ & $0.036\pm0.006$  & $0.162\pm0.027$ \\
			$\textbf{8.0}$ & $\textbf{72.47}\pm2.34$ & $\textbf{0.027}\pm0.011$  & $\textbf{0.127}\pm0.093$ \\
			8.5 & $72.22\pm2.23$ & $0.032\pm0.008$  & $0.177\pm0.067$ \\
			\bottomrule
	\end{tabular}}
\end{table}

\begin{figure}[t]
	\begin{center}
		\includegraphics[width=\linewidth]{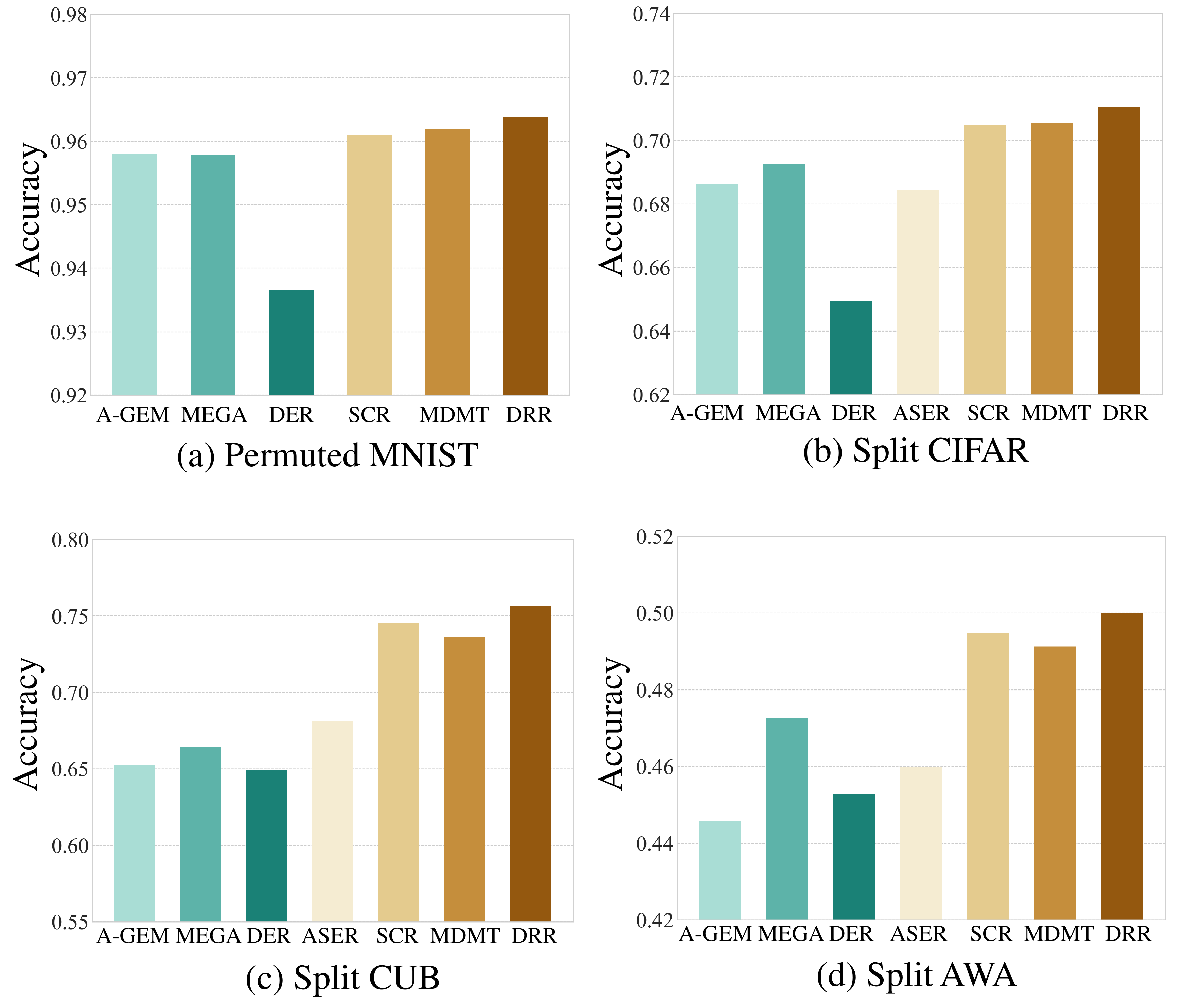}
    \vspace{-20px}
		\caption{Average accuracy of all tasks when they just finish their training.}
		\label{fig:lca}
	\end{center}
\end{figure}

\begin{figure}[t]
	\begin{center}
		\includegraphics[width=\linewidth]{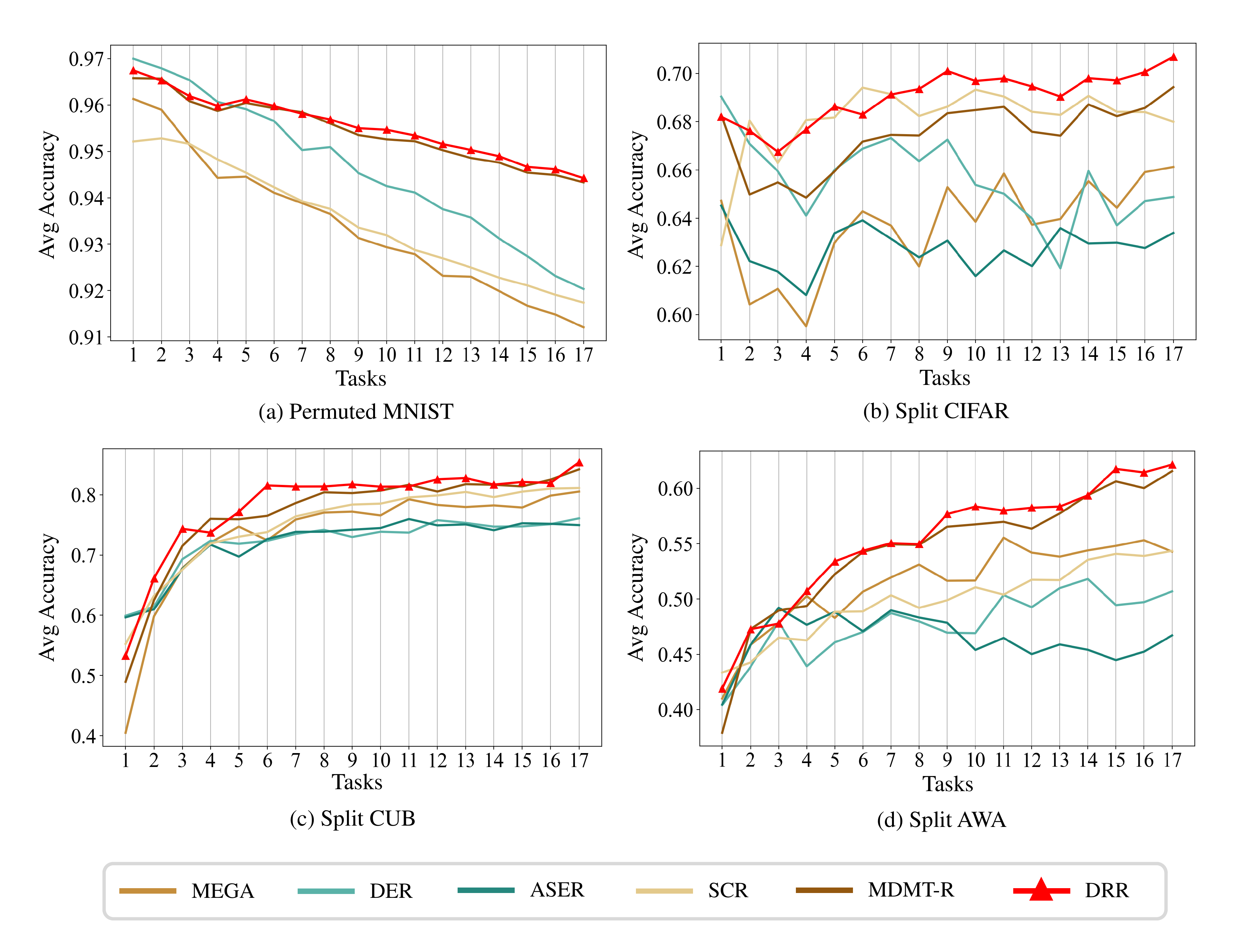}
    \vspace{-20px}
		\caption{
			Average accuracy trend (from $A_1$ to $A_T$) on four datasets.
		}
		\label{fig:acc}
	\end{center}
\end{figure}

\begin{table}[ht]
	\centering
	\caption{Sampling comparisons on Split CIFAR.}
	\label{tab:sample}
	\resizebox{\linewidth}{!}{
		\begin{tabular}{ccccc}
			\toprule
			Methods & {$A_{{T}}(\%)$} & {$F_{{T}}$}  & {$LTR$} & Time(ms)\\
			\midrule
			Ring buffer & $66.38\pm1.63$ & $0.052\pm0.006$ &   $0.377\pm0.076$ & ${27.53}$ \\
			MoF & $66.58\pm1.75$ & $0.053\pm0.010$ & ${0.359}\pm0.106$ & 32.62\\
			GSS & $62.06\pm3.58$ & $0.115\pm0.021$  & $0.912\pm0.183$ & 652.88\\
			ASER & $65.53\pm1.89$ & $0.072\pm0.007$  & $0.544\pm0.133$ & 44.13 \\ 
			\midrule
			DRR & ${69.37}\pm1.09$ & ${0.037}\pm0.007$ & ${0.253}\pm0.049$ &28.82\\
			
			\bottomrule
	\end{tabular}}
\end{table}


\begin{figure*}[t]
	\begin{center}
		\includegraphics[width=1\linewidth]{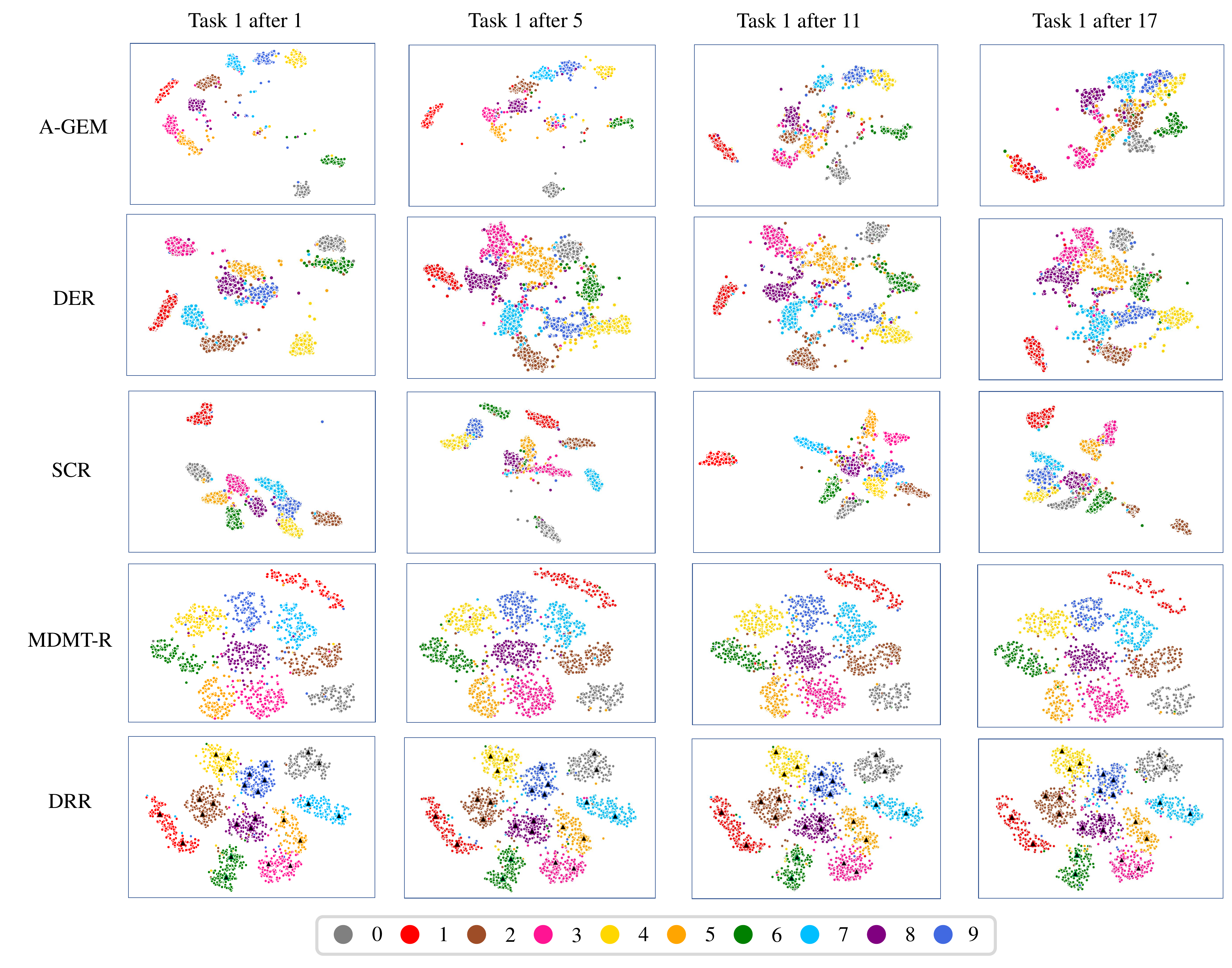}
    \vspace{-25px}
		\caption{The t-SNE results of the features from task 1 on Permuted MNIST after the lifelong learning on tasks 1, 9 and 17.
		}
		\label{fig:mnist_tsne}
	\end{center}
\end{figure*}

\subsection{Compared Methods}

We compare our DRR method with the following methods:

\noindent
 \textit{Regularization-based methods}. EWC~\cite{kirkpatrick2017overcoming}, MAS~\cite{aljundi2018memory}, RWalk~\cite{chaudhry2018riemannian} and SI~\cite{zenke2017continual} are regularization-based methods that set extra regularization terms to prevent forgetting. EWC and RWalk evaluate a Fisher information matrix to record the important parameters of old tasks. MAS adds perturbation to parameters to evaluate the importance of each neuron. SI is similar to MAS but watches the change of loss value.
\\
  \textbf{Reherasal-based methods}. GEM~\cite{lopez2017gradient}, MER~\cite{riemer2018learning}, ER~\cite{Chaudhry2019}, A-GEM~\cite{AGEM}, MEGA~\cite{guo2019learning}, DER~\cite{DER}, SCR~\cite{SCR}, $\epsilon$-SOFT-GEM~\cite{hu2020gradient}, ASER~\cite{ASER} and MDMT-R~\cite{lyu2021multi} are rehearsal-based methods that store and retrain on old data. GEM and A-GEM focus on finding the optimal gradient that saves the old tasks from being corrupt.
  MER combines meta-learning and experience replay to evaluate the importance of each neuron.
  ER concatenates the memory and current mini-batch and trains them together. 
  MEGA proposes to balance the loss via optimization.
  DER stores the logits or features of memory data to reduce catastrophic forgetting.
  SCR constructs contrastive loss and emphasizes reducing the negative effects of domain drift by boundaries between classes.
  $\epsilon$-SOFT-GEM adds a soft constraint to A-GEM to balance old and new knowledge.
  ASER stores samples near the classification boundary by calculating the adversarial shapely value.
  MDMT-R proposes to use a margins on softmax to control the decision boundaries.
  \\
 \textbf{VAN} means continual fine-tuning without a forgetting-reducing strategy. This is set as a lower bound of compared methods.
 We also evaluate the joint training (\textbf{JOINT}) on all datasets with different classifiers together. This is set as an upper bound of compared methods.


\subsection{Major Results}

As shown in Tables~\ref{tab:sota_mnist}, \ref{tab:sota_cifar}, \ref{tab:sota_cub} and \ref{tab:sota_awa}, we show the experimental results on four datasets including Permuted MNIST, Split CIFAR, Split CUB and Split AWA by three evaluated metrics including average accuracy $A_T$, forgetting measure $F_T$ and long-term remembering LTR.
For each metric, we have several observations on four datasets.





\textit{Comparisons on $A_T$}:
The average accuracy $A_T$ represents the final performance on each training task at the end time.
First, VAN, serving as the lower bound, obtains the worst performance, which means that the naive fine-tuning can hardly deal with catastrophic forgetting.
Second, joint training has the best performance because no forgetting occurs in this situation except on Split CUB. This is because CUB has classes that have large connections.
Thus, it is necessary to design continual learning in online incremental scenarios.
Third, we find that regularization-based methods still have a large performance gap compared to rehearsal-based methods, which means that the usage of memory is effective to reduce catastrophic forgetting in continual learning.
This also indicates that it is meaningful to suppress the domain drift in rehearsal-based continual learning.
Fourth, for $A_T$, our DRR method shows its superiority on all four datasets. 
This means less forgetting of old tasks and better learning of new tasks by reducing unpredictable continual domain drift in OCL.


\textit{Comparisons on $F_T$}:
$F_T$ evaluates the fine-grained batch-level forgetting, \ie, performance drop, but never cares about the accuracy value itself. 
Therefore, the forgetting measure depends on both the start and the end accuracy of each task.
As expected, VAN has the worst forgetting measure, because no strategy is used to reduce catastrophic forgetting.
Another observation is that rehearsal-based methods such as GEM and A-GEM have less forgetting than regularization-based methods.
On the four datasets, the proposed DRR obtains the best forgetting measure, which shows its effectiveness.
For example, on Split CIFAR, DRR achieves $0.03$ $F_T$ while the previous SOTA is MDMT-R ($0.04$).
Compared to MDMT-R, the proposed DRR achieves better performance in terms of both $A_T$ and $F_T$, which means better stability and plasticity.

\textit{Comparisons on $LTR$}:
LTR evaluates long-term remembering, and we use this metric to evaluate the influence of reducing drift.
On Permuted MNIST and Split CIFAR, DRR gets the best LTR ($0.243$ and $0.127$), and comparable LTR on Split CUB ($0.009$) and Split AWA ($0.009$).
Specifically, MDMT-R, gets the best LTR on Split AWA and DRR is only next to MDMT-R.
For Split CUB, the best method for LTR is ER, and we think this is because the dataset CUB has less discriminative on multiple bird classes, which means less impact of CML and ED losses because of similar representations.
For Split AWA, the dataset contains overlapped classes between different tasks, which may also reduce the degree of forgetting.


\subsection{Learning Trend Analysis}

In Fig.~\ref{fig:lca}, we show the average accuracy of all tasks when they just finish their training. 
The value represents the ability of learning new tasks, and the larger the better.
Obviously, the proposed DRR achieves the best value on the four datasets.
This means that DRR is able to reduce the domain drift influence on new tasks' training.
We observe that SCR obtains comparable results because it also considers reducing the negative effects of domain drift.
However, SCR ignores the quality of stored data just like MDMT-R, and may still suffer much from continual domain drift.

In Fig.~\ref{fig:acc}, we show the accuracy trends in the continual process (from $A_1$ to $A_T$).
Each value on a line means the average accuracy of seen tasks so far.
As observed from these results, the proposed DRR obtains robust results along the continual training process on the four datasets, which also indicates the better performance of the DRR.

\subsection{Ablation Study}

As shown in Table~\ref{tab:abl}, we analyze the importance of margins and CDL on Split CIFAR.
The first row is the results with only vanilla contrastive loss, while $m_\text{t}=0$ and $m_\text{c}=0$ means the Cross-Task Loss.
We have the following observations.
First, both CDL and CML improve the metric of average accuracy.
By adding both the CDL and CML, a dramatic performance improvement can be observed, which means that the two loss functions can clearly reduce catastrophic forgetting.
In Table~\ref{tab:abl2}, we then show the appropriate centroid threshold on Split CIFAR. 
We select a feasible threshold via grid searching. 
The results show that smaller or larger thresholds will bring performance drops on the three metrics.

\subsection{Comparison on Sampling Methods}

In Table~\ref{tab:sample}, we show the comparison of our centroid-based online selection with some other sampling methods including ring buffer, MoF~\cite{Rebuffi2016}, GSS(GSS)~\cite{gradient} and ASER~\cite{ASER}.
Note that ring buffer means random sampling.
For a fair comparison, we directly retrain stored memory of old tasks and only use different sampling strategies.
Our centroid-based online selection strategy outperforms all other compared sampling methods and with nearly $3\%$ improvement than them in terms of final average accuracy.
Moreover, our strategy is time-efficiency and uses close time for sampling.
The results show the proposed centroid-based online selection strategy is effective and efficient in rehearsal-based continual learning

\subsection{Visualization Analysis}

In this subsection, we show some visualizations of continual domain drift via t-distributed Stochastic Neighbor Embedding (t-SNE)~\cite{maaten2008visualizing} on Permuted MNIST.
As shown in Fig.~\ref{fig:mnist_tsne}, we show the continual domain drift of task 1 after the model trained on tasks 1, 5, 11 and 17, respectively.
Other methods including A-GEM, DER and SCR cannot reduce the continual domain drift at all, which results in serious forgetting.
The proposed DRR method can clearly reduce the continual domain drift.
By storing samples using centroids, our DRR can further reduce domain drift than MDMT-R. 
For example, evaluating task 1 after 17, MDMT-R has a diffused distribution compared to task 1 after 1. This phenomenon is because MDMT-R does not store representative samples for old tasks.
In comparison, our DRR can keep the distribution well even after 17 tasks.

\subsection{Computational Cost and Memory Complexity}

In Table~\ref{tab:cost}, we show the training computational cost and memory complexity on a single RTX 3090Ti GPU card.
We observe that the training time of the proposed DRR is comparable with other methods.
The memory cost of the training procedure is on the right of Table~\ref{tab:cost}.
Compared to the previous methods, the memory cost of the proposed method has a slight increase, $(M+C)\tilde{H}$, which arises from the stored features of old tasks.
However, the increase is very small compared to the total memory cost because in most situations the feature has a smaller size than an image.

\begin{table}[t]
	\centering
	\caption{Computational cost and memory complexity. The memory cost uses several parameters: (1) the number of tasks $T$; (2) the total number of parameters $P$; (3) the size of the mini-batch $B$; (4) the total size of the network hidden state $H$ (assuming all methods use the same architecture); (5) the size of the memory $M$ per task; (6) the size of the hidden state $\tilde{H}$ \wrt the latent representation; (7) the size of the logits $L$; (8) the number of data augmentation $A$. (9) the number of centroid $C$.}
	\label{tab:cost}
	\resizebox{\linewidth}{!}{
		\begin{tabular}{l|cccc|l}
			\toprule
			\multirow{2}{*}{Method}  & \multicolumn{4}{c|}{\textbf{Training time (ms/batch)}} & \multicolumn{1}{c}{\textbf{Memory}}  \\
			\cline{2-6}
			& MNIST& CIFAR & CUB & AWA & \multicolumn{1}{c}{Training} \\
			\midrule
			A-GEM 	& 2.27 & 19.96 & 60.41 & 59.47 & $2P+(B+M)H$  \\
			MEGA 	& 2.54 & 22.36 & 62.42 & 57.65 & $2P+(B+M)H$  \\
			DER 	& 25.69 & 59.91 & 109.05 & 79.90 & $2P+(B+M)H+ML$ \\
			ASER 	& - & 64.23 & 64.23 & 60.57 & $2P+(B+M)H$  \\
			SCR 	& 6.76 & 22.04 & 198.61 & 200.36 & $2P+(B+M+A)H$  \\
			MDMT-R 	& 5.98 & 49.85 & 123.29 & 94.88 & $2P+(B+M)H+M\tilde{H}$  \\
			DRR & 7.34 & 51.55 & 124.13 & 99.67 & $2P+(B+M)H+(M+C)\tilde{H}$ \\
			\bottomrule
	\end{tabular}}
\end{table}

\section{Conclusion and Future Work}
In this paper, we address catastrophic forgetting, a major challenge of OCL study, by considering the continual domain drift of old tasks in the training sequence. 
In rehearsal-based OCL, the continual domain drift is given rise from the imbalance of stored old tasks' data and massive new tasks' data.
We propose an effective Drift-Reducing Rehearsal (DRR) method, which effectively anchors the domain of old tasks and makes all tasks perceive each other. 
First, we store samples in the memory buffer of rehearsal by measuring the distance from the sample to constructed centroids.
Then, we propose a new cross-task contrastive margin loss to encourage intra-class and intra-task compactness, and inter-class and inter-task discrepancy.
Finally, we present an optional centroid distillation loss to mitigate the continual domain drift.
We evaluate the proposed DRR method on four OCL benchmark datasets. 
Extensive experiments show the superiority of our approach over SOTA methods. 

Although the method proposed in this paper has achieved certain effectiveness, we do not have a way to constrain the exact direction of domain drift, making the direction of drift easily uncontrollable after training on a large number of tasks. In the future, we plan to research modeling the direction of domain drift and controlling the orthogonality of multi-task directions to ensure better OCL effectiveness.

\section*{Acknowledgments}

This work was jointly supported National Science and Technology Major Project (2022ZD0117901), National Natural Science Foundation of China (Nos. 62373355, 62236010, 62276182, 61876220), and Postdoctoral Fellowship Program of CPSF under Grant Number (GZC20232993).
\ifCLASSOPTIONcaptionsoff
  \newpage
\fi



\bibliographystyle{IEEEtran}
\bibliography{ref}

\end{document}